\documentclass[accepted]{melba}

\usepackage{mwe} 

\usepackage{amsmath,amsfonts}

\usepackage{array}
\usepackage{algorithm}
\usepackage{algpseudocode}
\usepackage{diagbox}
\usepackage{longtable}

\usepackage{graphicx}
\usepackage{subcaption}

\usepackage[percent]{overpic}

\melbaid{2026:013}  
\doi{10.59275/j.melba.2026-f34b}
\melbaauthors{Fajardo-Rojas, Hall, Cromb, Rutherford, Story, Robinson, and Hutter}  
\email{diego.fajardo\_rojas@kcl.ac.uk}
\volume{2026}
\firstpageno{252}  
\melbayear{2026}  
\datesubmitted{2026-02-28}  
\datepublished{2026-06-05}  

\ShortHeadings{Predicting gestational age at birth from fetal MRI}{Fajardo-Rojas, Hall, Cromb, Rutherford, Story, Robinson, and Hutter}

\title{Predicting gestational age at birth \\ in the context of preterm birth from multi-modal fetal MRI}

\author{
	\firstname Diego \surname Fajardo-Rojas\aff{1,2}\orcid{0009-0001-8080-9226},
	\firstname Megan \surname Hall\aff{1,3}\orcid{0000-0003-2756-6365},
    \firstname Daniel \surname Cromb\aff{1}\orcid{0000-0002-9814-8841},
    \firstname Mary A. \surname Rutherford\aff{1}\orcid{0000-0003-3361-1337},
    \firstname Lisa \surname Story\aff{1,3}\orcid{0000-0001-9328-9592},
    \firstname Emma C. \surname Robinson\aff{2}\orcid{0000-0002-7886-3426},
    \firstname Jana \surname Hutter\aff{1,4}\orcid{0000-0003-3476-3500},
}
\affiliations{
	\num 1 \addr Early Life Imaging department, King's College London, London, UK \\
	\num 2 \addr Biomedical Computing department, King's College London, London, UK \\
	\num 3 \addr Women's Health department, School of Life Course and Population Sciences, King's College London, London, UK \\
    \num 4 \addr Institute for Information Processing, Leibniz University Hannover, Hannover, Germany \\
}

\abstract{
	Preterm birth is associated with significant mortality and a risk for lifelong morbidity. The complex multifactorial aetiology hampers accurate prediction and thus optimal care. A pipeline consisting of bespoke machine learning methods for data imputation, feature selection, and regression models to predict gestational age (GA) at birth was developed and evaluated from comprehensive multi-modal morphological and functional fetal MRI data from 333 control cases and 93 preterm birth cases. The GA at birth predictions were classified into term and preterm categories and their accuracy, sensitivity, and specificity were reported. An ablation study was performed to further validate the design of the pipeline. Performance was evaluated using stratified 10-fold cross-validation. The pipeline achieves an $R^2$ score of 0.13 and a mean absolute error of 2.74 weeks. It also achieves a 0.77 accuracy, 0.59 sensitivity, and 0.82 specificity across folds. The predominant features selected by the pipeline include cervical length and statistics derived from placental T2* values. The confluence of fast, motion-robust and multi-modal fetal MRI techniques and machine learning prediction allowed the prediction of the gestation at birth. This information is essential for any pregnancy. To the best of our knowledge, preterm birth had only been addressed as a classification problem in the literature. Therefore, this work provides a proof of concept. Future work will increase the cohort size to allow for finer stratification within the preterm birth cohort. 
	Our code is available at~\url{https://github.com/dfajardorojas/ml-for-preterm-birth-}.}


\begin{document}

\twocolumn[\maketitle]

\section{Introduction}
Preterm birth is defined as a live birth before 37 completed weeks of gestation \citep{WHO}. It is estimated that every year 13.4 million babies are born prematurely, corresponding to a global preterm rate of around 9.9\% \citep{ohuma}. Prematurity is the leading cause of mortality among children under 5 years accounting for 17.7\% of the 5.3 million yearly deaths in this age group \citep{perin}. Complications associated with preterm birth are also the leading cause of neonatal mortality, accounting for 36\% of these deaths \citep{perin}. The chances of survival of preterm babies are directly related to their gestational age (GA) at birth, with survival chances increasing from less than 18\% for babies born at 22 weeks to over 95\% for babies born at 29 weeks or later \citep{ancel, santhakumaran, bell}. Despite  advances in perinatal and neonatal care \citep{ancel, santhakumaran, bell, blencowe, cheong, boland} survival critically depends on every additional week in-utero. 

A continuous rise in survival rate has not translated into a decrease of the short- or long-term morbidity associated with preterm birth \citep{cheong, boland, allen_outcomes}. Short-term outcomes of premature birth include infections, bronchopulmonary dysplasia, retinopathy, necrotising enterocolitis, and brain disorders \citep{Costeloe2012ShortTO}. Long-term consequences include an increased risk of neuropsychiatric disorders such as psychosis, neurodevelopmental disabilities such as cerebral palsy and neuromotor dysfunction, adverse sensory outcomes such as hearing and visual impairment, as well as disabilities encompassing learning, cognition, and behaviour \citep{vanes, allen_outcomes, imad}. Similar to mortality rates, the incidence and severity of short- and long-term consequences of preterm birth are inversely related to GA at birth \citep{Costeloe2012ShortTO, moore, moster}. GA at birth is also correlated to social aspects later in life such as income and education level \citep{moster}.

Reducing the incidence of preterm birth and the impact of its consequences would not only alleviate the burden on individual patients and their families, but also on entire healthcare systems, since the lifetime cost of preterm births in the USA (in 2016) was estimated to be \$25.2 billion \citep{waitzman}.  Unsurprisingly, a review of the literature on the economic consequences of preterm birth found a prevailing inverse relation between economic costs and GA at birth, regardless of methodology, date, or country of publication \citep{petrou}. 

Preterm birth is classified into three subcategories: extremely preterm (less than 28 weeks), very preterm (28 to 32 weeks), and late preterm (32 to 37 weeks) \citep{WHO}, with further categorisation by clinical presentation: medically induced (or iatrogenic) and spontaneous \citep{moutquin}. While maternal and fetal indicators for iatrogenic preterm birth are well characterised and include conditions such as pre-eclampsia and fetal growth restriction (associated with $30.1\%$ of cases) \citep{goldenberg, morken}, the aetiologies underlying spontaneous preterm birth are complex, varied, and poorly understood \citep{frey}. Causes include\textemdash but are not restricted to\textemdash infection or inflammation, vascular disease (leading to uterine ischaemia), uterine overdistention, and cervical injury. The latter can be a consequence of LLETZ procedures, cervical cone biopsies for abnormal smear tests, and injuries resulting from emergency C-sections in previous pregnancies \citep{goldenberg, suff_csections}. However, definitive causes are registered for only 50\% \citep{menon, muglia} of cases. As such, spontaneous preterm birth should more broadly be considered a syndrome resulting from multiple intricate causes \citep{goldenberg, romero_parturition}.  

Despite this complexity, several risk factors have been identified \citep{goldenberg, frey, cobo} (see Table \ref{tab:risk_factors}) and are useful, both to provide insights and to help identify at-risk women. The wide variety of factors thereby matches the aetiological complexity of preterm birth. Even within the same clinical subtype, some factors can have opposite effects. For example, low maternal body mass index (BMI) is a risk factor for fetal growth restriction but protective against preeclampsia, whereas these roles are reversed for maternal obesity \citep{kramer}.

\begin{table}[!ht]
\centering
\caption{Most common risk factors for preterm birth \citep{goldenberg, frey, cobo}.}
\label{tab:risk_factors}
\begin{tabular}{|p{6.5cm}|}
\hline
\textbf{Risk Factors for Preterm Birth.} \\
\hline
African-American ethnicity \\
Depression \\
Family history of preterm birth \\
History of cervical excision \\
Infections (genitourinary or extragenital) \\
Low educational attainment \\
Low socio-economic status \\
Maternal age (low and high) \\
Maternal body mass index (low and high) \\
Multiple gestation (twins, triplets, etc) \\
Periodontal disease \\
Prior preterm birth \\
Stress \\
Stillbirth or induced abortion history \\
Tobacco use \\
Use of assisted reproductive technologies \\
Uterine anomalies \\
\hline
\end{tabular}
\end{table}

Currently there are three leading indicators used in clinical practice to identify women at high risk. The strongest predictor is a history of previous preterm birth or cervical surgery or injury (32\% chance of recurrent preterm birth) \citep{suff, goldenberg}. The other two biomarkers are mid-trimester cervical length below 25mm \citep{suff, romero_one}, measured via vaginal ultrasonography; and the presence of more than 50ng/mL fetal fibronectin, a glycoprotein usually absent in cervicovaginal fluid from 18 weeks of gestation and an indicator of choriodecidual disruption. The absence of any of these factors suggests the likelihood of delivering within the following 7 days is only around 1\% \citep{goldenberg, suff}. These factors have been combined within clinical practice to improve their predictive capabilities \citep{quipp, quipp_eval}. In other analyses, the combination of these predictors also reduced the average cost of high-risk pregnancies \citep{desplanches,vanbaaren}.

Imaging modalities beyond ultrasound are not currently incorporated into routine preterm birth risk assessment. Magnetic Resonance Imaging (MRI) is an imaging modality with good potential to investigate preterm birth. Fetal MRI can be used as a complementary modality to the commonly used ultrasound screening due to its higher resolution, operator independence and suitability for use on women with a higher BMI. It is also non-invasive with no evidence indicating any risk to the fetus or mother \citep{tocchio, lum, ray}. Another key advantage of MRI is that it offers  multiple complementary contrasts that can support comprehensive functional evaluation of fetal and maternal tissues \citep{story_antenatal}. Available contrasts include T2-weighted anatomical imaging, T2* relaxometry (which provides an indirect measure of oxygenation \citep{sorensen}), diffusion MRI (which can quantify alterations in tissue microstructure \citep{avena-zampieri, lee, slator, kristi}), flow measurements and T2 relaxometry. Past studies have largely focused on individual organs such as investigating changes to lung \citep{story_lungs}, thymus volumes \citep{story_thymus}, or assessing placental microstructure by measuring T2* and ADC values \citep{hutter}. One study measured umbilical vein T2 values as a potential marker of intrauterine growth restriction \citep{zhu}.

\section{Related Works}

While predictive machine learning (ML) models have enjoyed an ever-increasing popularity, preterm birth has only been addressed as a classification problem. Models based on electronic health records, uterine electromyography and transvaginal ultrasound \citep{wlodarczyk_review} reported accuracies of approximately $0.77$ \citep{esty, wlodarczyk_unet, prema}, while studies based on electrohysterography reported values above $0.94$ \citep{chen, despotovic, sadi-ahmed}, with the latter, however, only including records of women with recorded contractile activity \citep{sadi-ahmed, fele-zorz}.

Machine learning applied to structural MR measurements has been successful at predicting GA at the time of the scan during pregnancy. For example, Convolutional Neural Networks trained on fetal brain MRI have been able to outperform current clinical methods to estimate GA at the time of scan \citep{kojita, shen}. \cite{namburete} managed to obtain a mean absolute error of 6.1 days by developing bespoke features from 3D ultrasound and using a regression forest for prediction. 

For this work, a stacking approach was chosen to predict GA at the time of birth. Stacking is an ensembling technique that consists of combining the predictions of individual base models by training a meta-model \citep{zhou}. Stacking was introduced by \cite{wolpert} to improve the predictions and generalisability of individual classification models. \cite{breimanstacking} showed that stacking was also suitable for regression problems, while \cite{ting} generalised the technique further by stacking three different types of base models and exploring different meta-models than the ones used in previous work. Ensemble methods such as stacking have the statistical advantage of reducing the risk of overfitting to the training data by taking into account the predictions of all the base models, as well as the representational advantage of expanding the space of available models by combining the base models into meta-models \citep{dietterich}.

In recent years, stacking has been successful at various tasks such as genomic prediction \citep{liang}, protein interactions prediction \citep{yi}, or prostate cancer detection \citep{wang}. These works take advantage of more recent ML learning models, e.g. \cite{yi} use Support Vector Machines and XGBoost models as part of their base models, while \cite{wang} explore using a Random Forest as their meta-model. 

The present study combines a uniquely rich MR data acquisition including both anatomical and functional scans of multiple fetal organs, and multimodal MRI of the placenta, with a ML pipeline based on stacking. To the best of our knowledge, this is the first work to leverage the advantages of stacking methods together with a comprehensive multi-modal data set to predict GA at birth.

\section{Methods}

\begin{figure*}[!h]
    \centering
    \includegraphics[width=14cm]{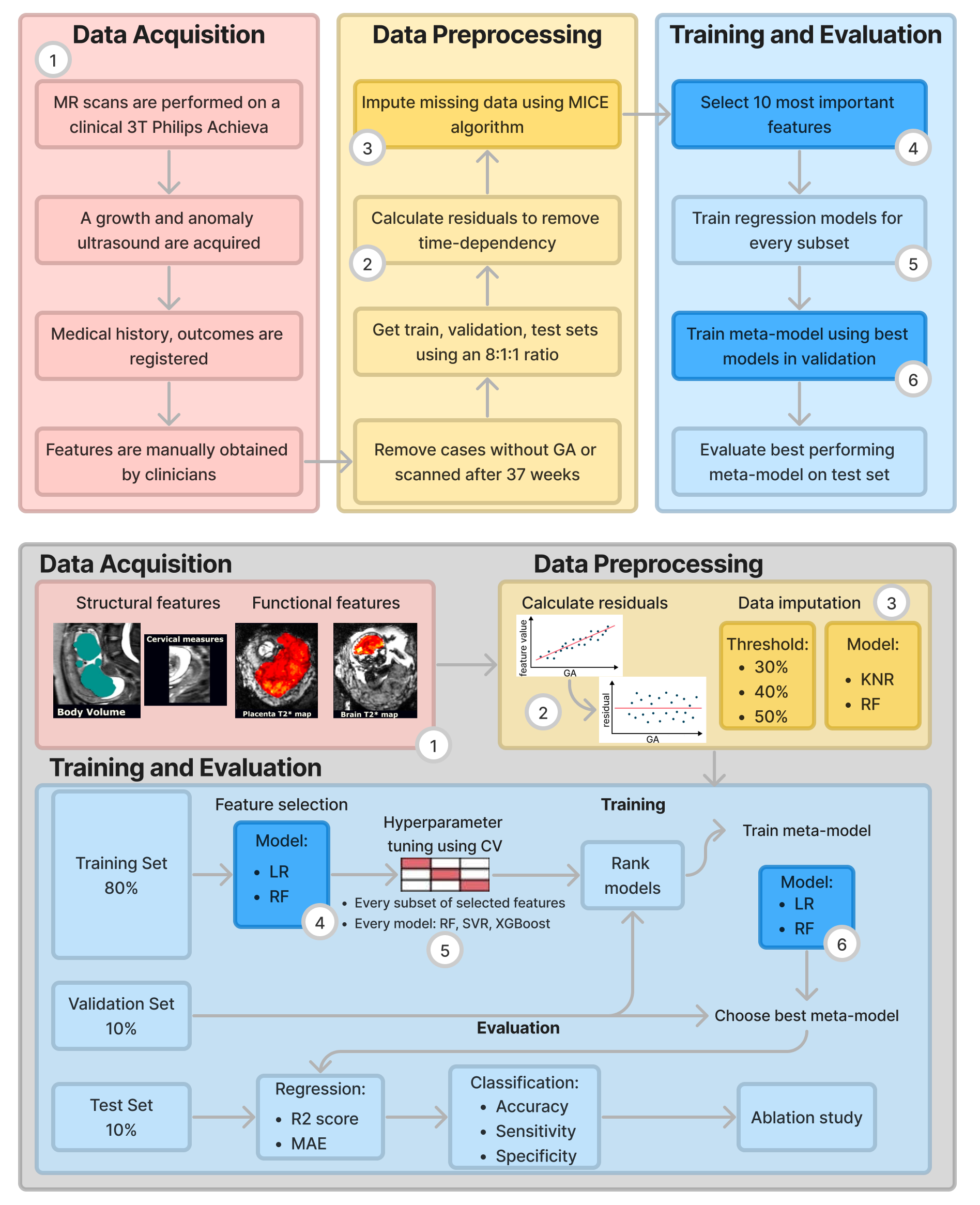}
    \caption{{\bf Schematic representation of the project pipeline.} The boxes in a darker shade denote steps of the pipeline where different design options were explored. The top part of the figure shows the flow of the pipeline with fixed design options, while the bottom part explicitly indicates the different design choices available for each step.} 
    \label{fig:pipeline}
\end{figure*}

This section contains a detailed outline of the development of the ML pipeline introduced in this work. The pipeline was designed to address the challenges presented by the data. These include: a large number of derived features relative to the number of training examples, data imbalance, and missing data. These problems were addressed through feature selection, balanced training, and feature imputation. Throughout the development of the pipeline, different design options were investigated including changing data threshold for imputation, and models for feature selection and regression. The end product is a meta-model where predictions are stacked to obtain a final predicted GA at birth. An ablation study, which investigates the impact of each component, is also described in detail. Fig \ref{fig:pipeline} illustrates the workflow of the project. The reader is invited to refer to it repeatedly to complement the description that follows.

\subsection{Data}\label{subsection:Data}
The data set used for this work comprises clinical records, MR data, and parameters manually extracted from ultrasound from 489 singleton pregnancies. From the 489 cases originally considered, 47 cases were excluded as they were lacking GA at delivery. We also removed 16 cases scanned after 37 weeks, since \textemdash in the context of predicting preterm birth\textemdash these would bias training of the model. This resulted in a final data set of 426 cases (see Appendix \ref{app:figure1}). 

Recruitment for all the considered studies was opportunistic, with two studies particularly recruiting women at high risk of preterm birth based on obstetric history, ultrasound and biomarker findings. However, the stated difficulty in accurately predicting preterm birth renders this task difficult, and as a result recruitment and thus the data set available is  biased towards term birth.


Data were split into 10 stratified folds of the same size, i.e. keeping an equal proportion of term and preterm birth cases in each set. These folds were used to obtain train, validation, and test sets with an 8:1:1 ratio across cross-validation iterations.

\subsection{Image Acquisition and Processing}

Imaging protocols were similar for each study: MR scans were performed on a clinical 3T Philips Achieva scanner between 15 and 40 weeks of gestation using a 32-channel cardiac coil (as is standard process for fetal imaging). All mothers were scanned in supine position. For maternal comfort, padding was provided, imaging time was limited to under 90 minutes, and there was frequent verbal interaction and monitoring of heart rate and blood pressure. The protocol included both anatomical T2-weighted imaging and functional MR sequences. For the work presented here only T2* relaxometry among the functional sequences was used.

Anatomical information was acquired with a 2D multi-slice Turbo-Spin-Echo sequence in four to ten planes covering the fetal brain and uterus. Next, to allow for image-based shimming on the 3T scanner, a map of the B0 field was obtained; then shimming was performed for the organ of interest. Afterwards, functional MRI of the entire uterus was performed in coronal orientation using free-breathing multi-echo Gradient Echo with Echo Planar read-out. 

In addition to the MRI, two ultrasound scans were performed: an anomaly scan (clinically performed between 19 and 21 weeks of gestation) and a second growth scan (including Doppler ultrasound) that was generally performed within one week of the MRI. In both cases, morphological measurements were manually extracted including abdominal and head circumference, bi-parietal diameter (i.e. the cross-sectional diameter of the skull), femur length, and expected weight. From the growth scan blood flow pulsatility indices were also estimated for the umbilical, uterine and mid-cerebral arteries. 

The obtained MR images were processed to obtain quantitative values. For the anatomical data, slice-to-volume reconstruction \citep{uus2020deformable}  and learning-based segmentation  \citep{uus2023bounti} were applied separately for the brain, body and the placenta. Then regional volumes were calculated. 

From the functional data, no motion correction was applied, since all echos for each slice were acquired within 200ms. Using the method described in \cite{hutter}, quantitative T2* values were obtained by fitting the signal of data from subsequent echo times for the entire uterine field of view (FOV). Values over 300ms were clipped to limit partial volume effects following common practise. Segmentation of the placenta, brain and lungs was performed manually. From these segmentations, regional volumes were calculated, as well as the mean, kurtosis, and skewness of their T2* distributions. These data acquisition steps are represented by index 1 in Fig \ref{fig:pipeline}.

\subsection{Summary of Derived Features}

In addition to imaging derived features, demographics, obstetric and medical history of the patients (including previous pregnancies, miscarriages and preterm births) were recorded as well as any relevant information from the current pregnancy such as diagnosis of pre-eclampsia, gestational diabetes, fetal growth restriction or any other fetal or maternal pathology. Finally, the outcomes of the pregnancy were obtained, including gestational age at birth, birth weight-centile and any occurrence of major complications.  Collectively the full set of features used by our models was summarised as follows (see Appendix \ref{app:features} for more details):
\begin{enumerate}
     \item \textbf{Clinical variables}: e.g. number of previous preterm deliveries, and maternal body mass index.
     \item \textbf{Structural MRI metrics}: describing sizes of structures e.g. volumes of different brain regions or bi-parietal diameter of the foetal head; these were extracted from both the anatomical and functional scans
     \item \textbf{Functional MRI metrics}: statistics derived from T2* distributions of the placenta, brain and lungs.
     \item \textbf{Ultrasound metrics}: from both anomaly and growth ultrasounds - manually extracted by a trained sonographer e.g. the fetal head circumference, femur length.
\end{enumerate}

\subsection{Feature cleaning}

Prior to training it was vital to address the confound effect of gestational age at scan, as well as address the impact of missing data.

\subsubsection{Deconfounding}  While GA at scan is a feature that would normally be available in a clinical setting, its impact on any learning model could lead to data leakage (e.g. by acting as a lower bound for GA at delivery). Moreover, as all features change dramatically with age \citep{story_lungs, story_brains, story_thymus, papageorghiou}, it is necessary to disentangle the dominant effect of GA from more subtle signatures that might robustly predict preterm birth. For these reasons, GA was linearly regressed from all features using the method of internally studentised residuals \citep{cook}. See index 2 in Fig \ref{fig:pipeline}.

\subsubsection{Data Imputation}
There was significant heterogeneity in the availability of features across the data set. Fetal and maternal motion, maternal discomfort, and clerical errors led to loss of data, with different features available for each of the 426 cases. For this reason, a regression-based approach to imputation, known as Multivariate imputation by chained equations (MICE)  \citep{azur}, was investigated (see Appendix \ref{app:mice}). Following guidance from the literature, ten iterations of the model were performed \citep{azur, jager}, with two different regression models: weighted $K$-Nearest Neighbours (KNR) \citep{bicego} and Random Forests (RF) \citep{breiman}. Both models were implemented in the standard way using Sci-kit Learn \citep{scikit-learn}. Imputation should not be applied to features with arbitrarily large amounts of missing data \citep{bertsimas, jager}. Thus the impact of discarding features with more than 30\%, 40\%, or 50\% missing values was investigated (see Appendix \ref{app:statistics} for the missing percentages of each feature).  Features with a greater percentage of missing values than the respective threshold were discarded. All remaining features were normalised (mean 0, std 1) afterwards. 

Data imputation corresponds to index 3 in Fig \ref{fig:pipeline}. The boxes corresponding to this step are emphasised by a darker shade to represent that different options were investigated as part of the pipeline design process. The top part of the figure shows the flow of the pipeline with fixed design choices (e.g. if the choice is made to investigate a pipeline using a RF within the MICE algorithm to impute features with less than 40\% of missing data). Conversely, the bottom part of the figure explicitly indicates the design choices that were investigated for this step.

\subsection{Training} 

Training was performed using a stacking approach in which a number of different classes of machine learning model were trained and these were ensembled together through the training of a meta model \citep{wolpert}. Base models consisted of: Random Forests (RF) \citep{breiman}, Support Vector Regression (SVR) \citep{smola}, and XGBoost \citep{XGBoost}. Each was chosen due to unique strengths: RF are interpretable and robust to overfitting \citep{gzar}; SVR are robust to outliers and well-suited to small data sets \citep{FernandezDelgado, kinaneva}; XGBoost offers state-of-the-art performance from sparse data sets \citep{XGBoost}. Importantly they are all capable of capturing non-linear relationships but approach regularisation in different ways \citep{FernandezDelgado}. This suggests that they will perform differently on boundary cases, to produce diverse predictions that could benefit from ensembling. 

Since a key challenge of training models on our data set has been the high number of features relative to examples (see Appendix \ref{app:features}), feature selection was also performed to discourage overfitting. Two simple models were explored: Linear Regression (LR) and Random Forests. For each model trained, 10 features were selected. These two different design options are indicated by the boxes with a darker shade with index 4 in Fig \ref{fig:pipeline}.

Models were trained using the Sci-kit learn framework, with hyperparameters (see Appendix \ref{app:hyperparameters}) optimised using 3-fold cross-validated grid search \citep{krstajic}. The metric used for optimisation was the coefficient of determination ($R^2$) \citep{casella}. Given fixed design choices on the previous steps, training was carried out every non-empty subset of the selected features. Since there are $1023$ non-empty subsets of the ten selected features and $3$ regression models, $3069$ different regression models were trained in total (index 5 in Fig \ref{fig:pipeline}). These were then composed via the training of a meta-model, for which two different methods were explored: Linear Regression and Random Forests (index 6 in Fig \ref{fig:pipeline}). Meta-models were trained on the $m$ best performing base models, as validated through their $R^2$ score on the validation set. The value of $m$ was also optimised using the validation set. 

The procedure described in this subsection was performed independently across 10 cross-validation iterations using the splits introduced in subsection \ref{subsection:Data}.

\subsection{Ablation Study}\label{subsection:ablation}
An ablation study was conducted to validate the design of the proposed pipeline, with results compared against the best performing meta-model. Due to the computational cost of the full pipeline, the study was performed on the cross-validation iteration with the highest performance on the test set. Since XGBoost models may be trained with incomplete data, and without variance normalisation of the features (since the base learners are decision trees) the first two experiments consist of a single XGBoost model trained on unnormalised data. All experiments are described as follows:

\begin{enumerate}
\item Out-of-the-box XGBoost: XGBoost without preprocessing.

\item XGBoost with deconfounding: one XGboost was trained after linear deconfounding of features. 

\item Imputation: all base predictive models were trained with deconfounding and imputation  (using the imputation approach used in the best meta-model), without performing any upsampling or feature selection. 

\item Correcting data imbalance: base models were trained with imputation and upsampling preterm cases in the training set, without performing any upsampling or feature selection. 

\item Feature selecting and upsampling: This equates to evaluating the best performing base model, obtained without ensembling.  

\item Meta-model without upsampling: the impact of upsampling in the whole pipeline was explored by turning it off. This is equivalent to the final meta-model without upsampling.

\item Meta-model: Reporting the performance of the proposed meta-model - obtained from the whole pipeline.  

\end{enumerate}

\subsection{Comparison with Existing Classification Studies}

As mentioned in the Introduction, preterm birth prediction has primarily been addressed as a classification problem. Although the primary objective of this work is to predict GA at birth as a regression task, these predictions can be classified into term or preterm, allowing comparison with previous classification-based approaches. A comparative analysis was performed to place the results of this work within the context of existing classification-based studies.

\section{Results}
\subsection{Data Exploration}\label{subsection:data_exploration}

\begin{figure*}[!h]
    \centering
    \includegraphics[width=13cm]{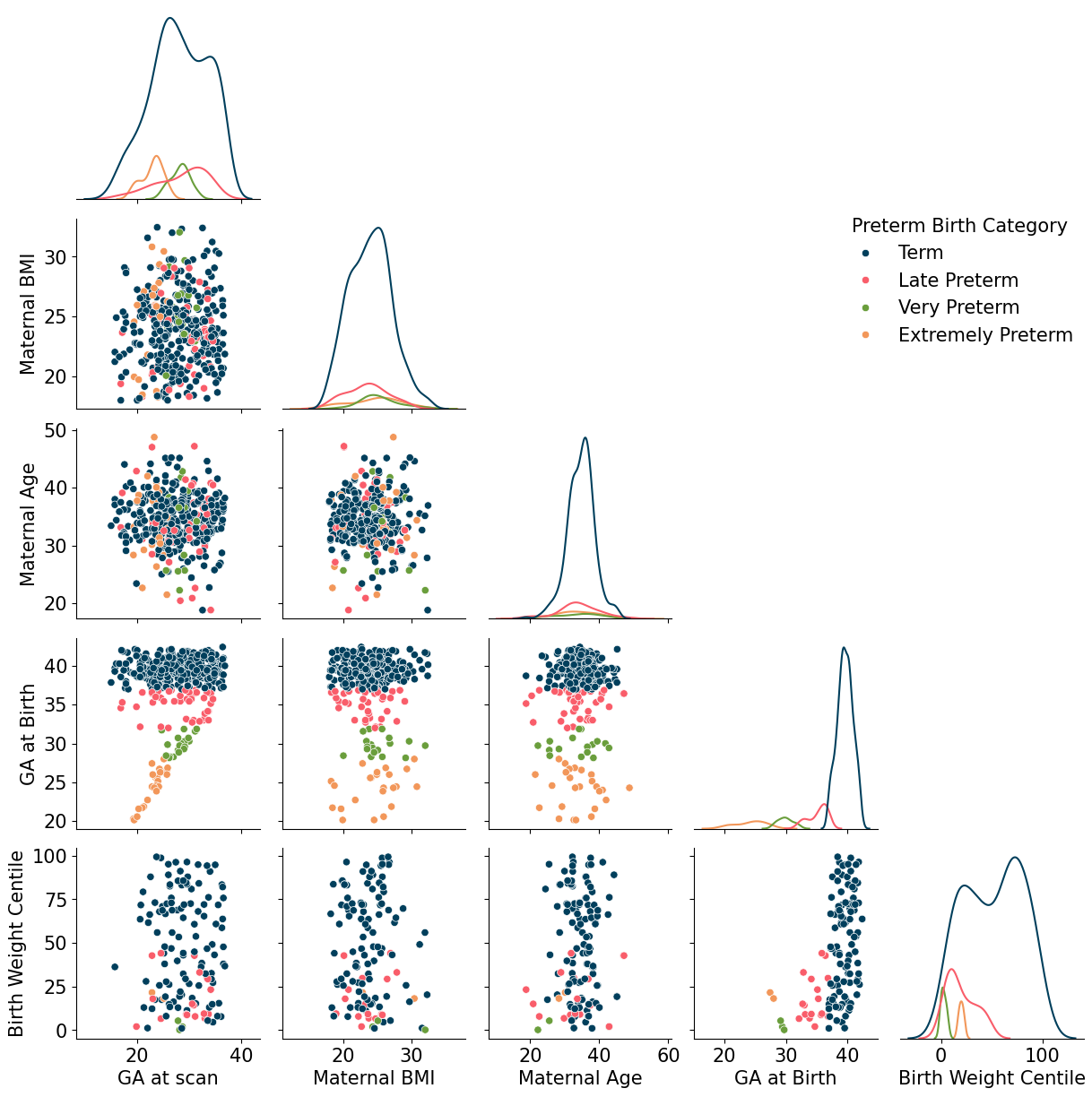}
    \caption{{\bf Data exploration.} Distributions of GA at scan, maternal BMI at scan, maternal age, GA at birth, and birth weight centile (diagonal panels), and their pairwise relationships (off-diagonal panels). Data are colour-coded by preterm birth category.}
    \label{fig:demographics4}
\end{figure*}

Table \ref{table:demographics} shows key demographics, clinical information, and outcomes, divided into preterm and control cohorts. Specifically, the data set consisted of 333 control cases and 93 preterm cases. The distribution of the data according to the four temporal categories was $78.2\%$ term, $10.8\%$ late preterm, $4.9\%$ very preterm, and $6.1\%$ extremely preterm. Fig \ref{fig:demographics4} shows the distribution of the five continuous features and outcomes included in Table \ref{table:demographics}, namely GA at scan, maternal BMI at scan, maternal age,  GA at birth, and birth weight centile. The pairwise relationship between these is also plotted. For a statistical summary of the data set see Appendix \ref{app:statistics}.

\begin{table*}[!ht]
    \centering
    \small
    \caption{\textbf{Key demographics, clinical information, and outcomes of participants.} Reported values are the mean $\pm$ SD in the case of continuous variables and percentages for discrete variables. The numbers in brackets are ranges.}
    \begin{tabular}{|l|c|c|}
    \hline
&\textbf{Preterm birth cohort} & \textbf{Control cohort} \\
    \hline
\multicolumn{3}{|c|}{\textbf{Current pregnancy}}\\    
    \textbf{Gestational age at scan [weeks]} & 26.88$\pm$4.40 [16.86,34.57] & 28.15$\pm$5.43 [15.00,36.86]     \\     \textbf{Maternal BMI at scan [kg/m2]} & 24.24$\pm$3.21 [18.26,32.05] & 24.04$\pm$2.94 [18.00,32.46]      \\
   \textbf{Maternal age at scan [years]} & 33.75$\pm$6.08 [18.82,48.74] & 34.75$\pm$3.84 [18.81,45.20]       \\
\hline
\multicolumn{3}{|c|}{\textbf{Obstetric history}}\\    
    \textbf{Previous preterm birth} & 13.98\% & 5.71\%     \\
\hline
\multicolumn{3}{|c|}{\textbf{Outcome}}\\    
    \textbf{Gestational age at birth} & 30.94$\pm$5.03 [20.14,36.86] & 39.67$\pm$1.24 [37.00, 42.43]     \\
    \textbf{Birth weight centile} & 35.34$\pm$32.75 [0.00,95.53] & 55.65$\pm$26.69 [0.00,99.97]     \\
    \textbf{Fetal sex} & 53.93\% female, 46.07\% male & 53.40\% female, 46.60\% male     \\
    \hline
    \end{tabular}
    \label{table:demographics}
\end{table*}

\subsection{Meta-model}\label{subsection:meta_model_results}

The best performing meta-model was obtained using the following settings in the pipeline. First, features with more than 50\% missing values were discarded. Then, features with 50\% or less missing values were imputed using the MICE algorithm with a KNR as its regression model and a RF was used for feature selection. After training RF, SVR, and XGBoost models with every non-empty subset of the selected features, the 18 models with the highest $R^{2}$ score on the validation set were used as input for a RF meta-model. In what follows, this meta-model will be referred to by abbreviating its components, i.e. 50-KNR-RF. 

\begin{figure}[!h]
  \centering

  \begin{overpic}[width=9.5cm]{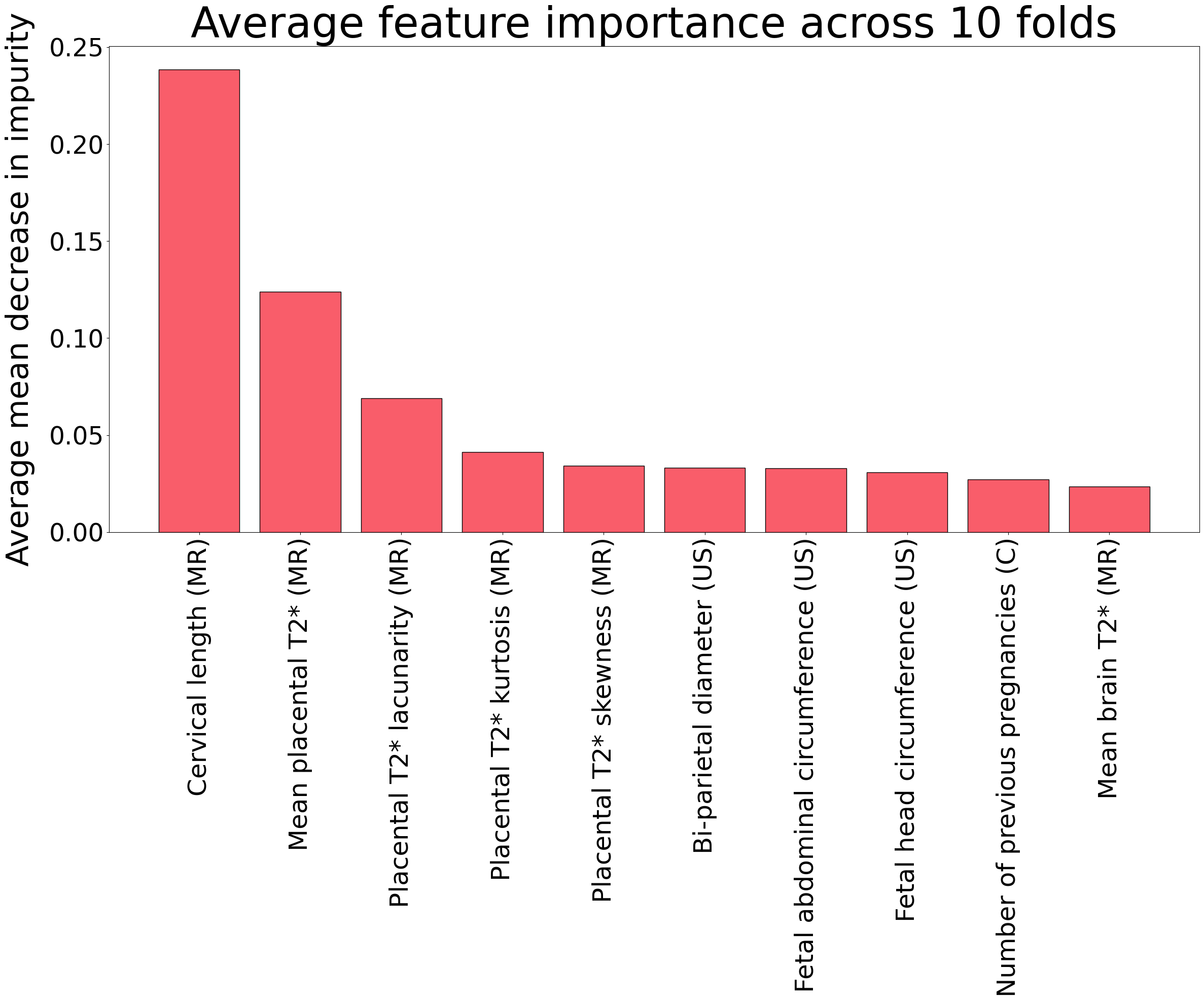}
    \put(4,82){\large\textbf{(a)}}
  \end{overpic}

  \vspace{1em}

  \begin{overpic}[width=9.5cm]{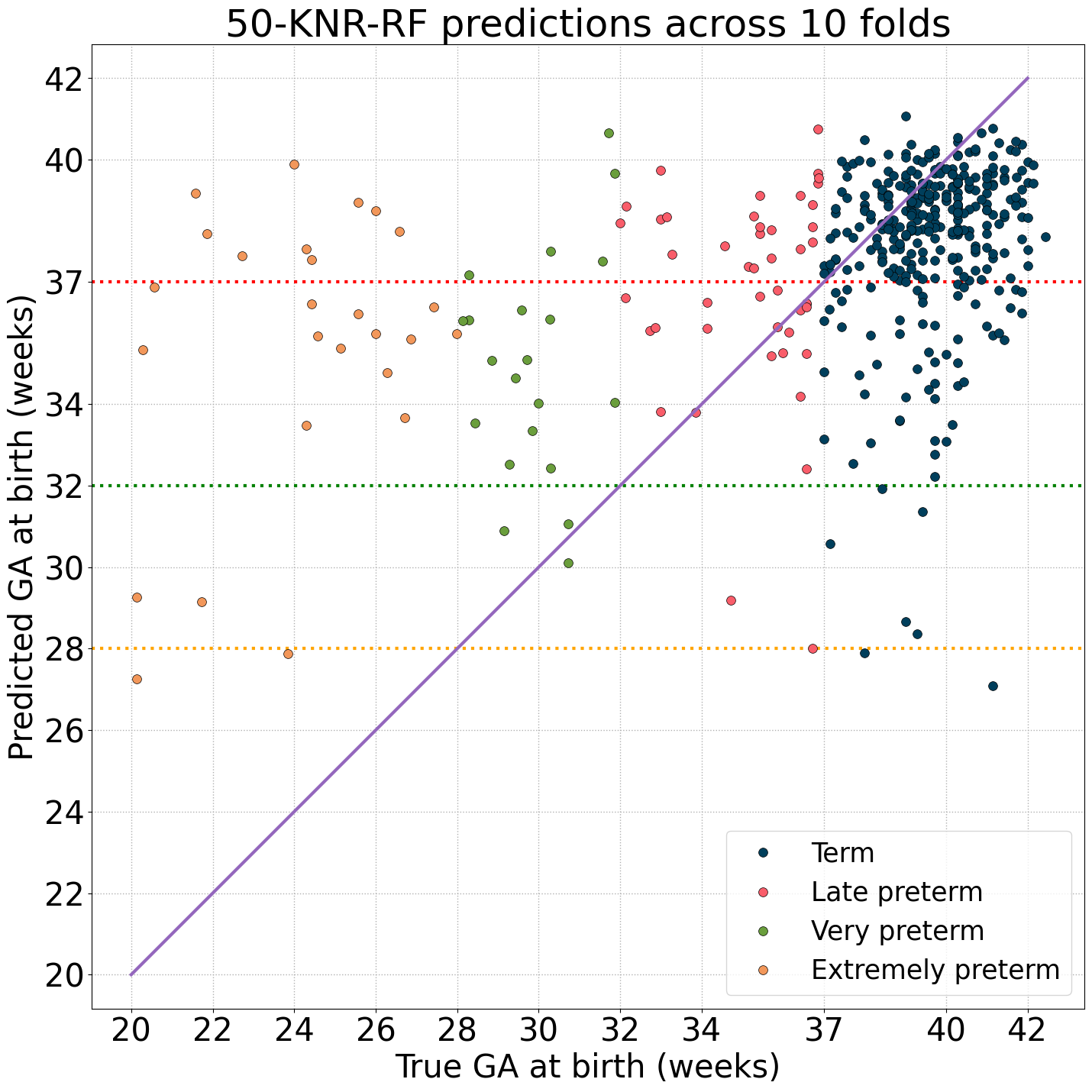}
    \put(4,97){\large\textbf{(b)}}
  \end{overpic}

  \caption{{\bf Feature importances and predictions of the meta-model.} (a) Average mean decrease in impurity for the ten highest-ranked features based on their average importance across cross-validation folds. (b) Predictions made by the meta-model 50-KNR-RF across cross-validation folds, colourised according to their true preterm temporal category.}
  \label{fig:results}
\end{figure}

Although different features were selected in each of the ten cross-validation iterations, cervical length, mean placental T2*, placental T2* lacunarity, and bi-parietal diameter were always selected. Fig. \ref{fig:results} (a) shows the average mean decrease in impurity for the ten features with the highest average importance across folds. This is the metric used by Random Forests to quantify the importance of each feature \citep{nembrini}.


For every iteration, the metrics used for evaluation were the $R^2$ score and the mean absolute error (MAE) measured in weeks. The cases were labeled as term ($\geq 37$ weeks) or preterm ($<37$ weeks), according to the GA predicted by the meta-model, and accuracy, sensitivity, and specificity were also reported. Table \ref{table:cross-val} shows the performance of the meta-model across the ten cross-validation folds. On average, the model achieved an $R^{2}$ score of $0.13$ and a MAE of $2.74$ weeks, as well as $0.77$ accuracy, $0.59$ sensitivity, and $0.82$ specificity. In eight of the ten folds, the MAE was less than 3 weeks. Sensitivity greater than $0.66$ was observed in six of the ten folds, although it remained below $0.5$ in the remaining four. On the other hand, specificity remained consistently high ($\geq 0.67$), showing a more reliable identification of term cases. The predictions made by 50-KNR-RF on all folds are depicted in Fig. \ref{fig:results} (b).

\begin{table*}[!ht]
    \centering
    \footnotesize
    \caption{{\bf Cross-validation results.}}
    \begin{tabular}{ |l|c|c|c|c|c| }
    \hline
   \multicolumn{1}{|c|}{ \textbf{Fold}} & \textbf{R}$\boldsymbol{^{2}}$ \textbf{score} & \textbf{MAE}  & \textbf{Accuracy}  & \textbf{Sensitivity}  & \textbf{Specificity} \\
    \hline
    Fold 1 &  0.05 &	2.70 	&	0.74 &	0.22 &	0.88  \\
    \hline
    Fold 2 &  0.15 &	2.62 &	 0.81 &	0.67 &	0.85 \\
    \hline
    Fold 3 &  -0.05 &	3.50 &	 0.63 &	0.33 &	0.71  \\
    \hline
    Fold 4 &  -0.22 &	3.80 &	 0.60 &	0.40 &	0.67	 \\
    \hline
    Fold 5 &  0.07 &	2.99 &  0.77 &	0.70 &	0.79  \\
    \hline
    Fold 6 &  0.42 &	2.50 &	 0.77 &	0.70 &	0.79  \\
    \hline
    Fold 7 &  0.08 &	2.64 &	0.76 &	0.89 &	0.73	 \\
    \hline
    Fold 8 &  0.06 &	2.17 &	 0.79 &	0.44 &	0.88	 \\
    \hline
    \textbf{Fold 9} &  \textbf{0.47} &	\textbf{2.19} &	 \textbf{0.83} &	\textbf{0.67} &	\textbf{0.88}	 \\
    \hline
    Fold 10 &  0.29 &	2.29 &	 0.98 &	0.89 &	1.00	 \\
   \hline
   \textbf{Average} &  \textbf{0.13} &	\textbf{2.74} &	 \textbf{0.77} &	\textbf{0.59} &	\textbf{0.82}	 \\
   \hline
    \end{tabular}
    \label{table:cross-val}
\end{table*}


To explore whether performance differed across clinically relevant obstetric subgroups, we evaluated the final meta-model stratified by parity and previous preterm birth history. The reported metrics correspond to the average performance across the ten cross-validation folds and are presented in Table \ref{tab:stratified_performance}. Performance in both parity subgroups was comparable to that observed in the full cohort, with similar MAE (2.75 weeks), accuracy (0.77), and sensitivity (0.59). This is also the case for the cohort of women without a previous preterm birth history. In contrast, the model performance worsened within the subgroup of women with a previous history of preterm birth, demonstrating reduced accuracy (0.66) and increased MAE (3.18 weeks), although sensitivity remained consistent with that of the full cohort (0.61).

\begin{table*}[htbp]
\centering
\caption{{\bf Model performance across 10 folds stratified by parity and previous preterm birth history (PBH).}}
\label{tab:stratified_performance}
\begin{tabular}{|l|c|c|c|c|c|c|}
\hline
\textbf{Group} & $\boldsymbol{n}$ & \textbf{R}$\boldsymbol{^{2}}$  \textbf{score} & \textbf{MAE} & \textbf{Accuracy} & \textbf{Sensitivity} & \textbf{Specificity} \\

\hline

\multicolumn{7}{|l|}{\textbf{Parity}}\\
\hline
Nulliparous women & 302 & 0.05 & 2.75 & 0.77 & 0.59 & 0.81 \\
Multiparous women & 124 & 0.24 & 2.73 & 0.77 & 0.59 & 0.84 \\
\hline

\multicolumn{7}{|l|}{\textbf{Previous preterm birth history}}\\
\hline
No previous PBH & 394 & 0.12 & 2.71 & 0.78 & 0.59 & 0.82 \\
Previous PBH & 32 & 0.01 & 3.18 & 0.66 & 0.61 & 0.68 \\

\hline

\multicolumn{7}{|l|}{\textbf{Full cohort}}\\
\hline
Full cohort & 426 & 0.13 & 2.74 & 0.77 & 0.59 & 0.82 \\
\hline
\end{tabular}
\end{table*}



\subsection{Ablation study}\label{subsection:ablation_study_results}

In line with the optimisation metric used during training, Fold 9 was selected for the ablation study, as it achieved the highest $R^2$ score among the cross-validation folds (Table \ref{table:cross-val}). The performance of each of the models in the ablation study is reported in Table \ref{table:ablation}. 

\begin{table*}[!ht]
    \centering
    \footnotesize
    \caption{{\bf Evaluation of the models in the ablation study.}}
    \begin{tabular}{ |l|c|c|c|c|c| }
    \hline
   \multicolumn{1}{|c|}{ \textbf{Model}} & \textbf{R}$\boldsymbol{^{2}}$ \textbf{score} & \textbf{MAE}  & \textbf{Accuracy}  & \textbf{Sensitivity}  & \textbf{Specificity} \\
    \hline
    1) Out-of-the-box XGBoost &  0.10 &	2.75 	&	0.71 &	0.56 &	0.76  \\
    \hline
     2) XGBoost with deconfounding &  0.03 &	2.96 &	 0.67 &	0.56 &	0.7	 \\
    \hline
    3) Imputation, RF &  0.18 &	2.89 &	 0.64 &	0.33 &	0.73	 \\
    \hline
    3) Imputation, SVR &  0.16 &	2.44 &	 0.83 &	0.22 &	1	 \\
    \hline
    3) Imputation, XGBoost &  0.28 &	2.66 &  0.69 &	0.33 &	0.79  \\
    \hline
    4) Correcting data imbalance, RF &  -0.29 &	4.09 &	 0.21 &	0.89 &	0.03  \\
    \hline
    4) Correcting data imbalance, SVR &  0.01 &	3.21 &	0.69 &	0.56 &	0.73	 \\
    \hline
    4) Correcting data imbalance, XGBoost &  0.27 &	2.74 &	 0.64 &	0.67 &	0.64	 \\
    \hline
    5) Feature selecting and upsampling, RF &  0.37 &	2.48 &	 0.76 &	0.67 &	0.79	 \\
    \hline
    6) Meta-model without upsampling & 0.39	& 2.38 &	0.79 &	0.44 &	0.88 \\
    \hline
    7) Meta-model, 50-KNR-RF & 0.47 & 2.19 &	0.83 &	0.67	&  0.88 \\
    \hline
    \end{tabular}
    \label{table:ablation}
\end{table*}

These results are helpful to understand the contributions and limitations of every element of the pipeline. In experiments 1) and 2) it can be seen that the out-of-the-box XGBoost model outperforms XGBoost trained after feature deconfounding, which is consistent with XGBoost’s predictive capabilities with minimal feature preprocessing \citep{XGBoost}. 

The sensitivity of all models trained without upsampling is less than $0.6$, while the sensitivity of all but one of the models trained with upsampling is greater than $0.66$. This suggests that upsampling was effective at addressing class imbalance. Applying feature selection in addition to upsampling results in a better overall performance than using upsampling alone: the RF obtained in experiment 5) achieves better regression and classification metrics than all models from previous experiments. It can also be noted that for individual models (i.e. experiments 3) against 4)) the increase in sensitivity is observed alongside a higher MAE. This trade-off was not observed for the final meta-model (i.e. experiments 6) against 7)) where the complete pipeline achieves a better performance across all metrics than its counterpart.

Lastly, the 50-KNR-RF meta-model achieved the best and most balanced performance, exhibiting the highest accuracy ($0.83$) along with high sensitivity ($0.67$) and specificity ($0.88$). While models like the SVR in experiment 3) and the RF in experiment 4) attained higher specificity and sensitivity, it can be seen that these models are highly biased towards predicting term and preterm cases respectively.

\subsection{Comparison with Existing Classification Studies}

Classifying the predictions of 50-KNR-RF into term or preterm allows for comparison with other models in the literature. As shown in Table~\ref{table:discussion}, the sensitivity achieved by 50-KNR-RF is lower than that reported by previous studies. However, such comparisons should be interpreted taking into account important methodological differences. \cite{wlodarczyk_unet} developed an automated method for cervical length and anterior cervical angle measurements from transvaginal ultrasound. Subsequently, they make preterm birth predictions based on these features, but explicitly use a balanced data set  with precomputed markers for their classification experiments instead of their original data set. \citep{esty} develop their models on very large population-level registry data sets (over 669,000 subjects) and include features such as obstetrical complications and registers of premature rupture of membranes\textemdash not only lead to preterm delivery \citep{goldenberg} but\textemdash under the framework of the present work would be considered a clinical outcome and would generally not be available at the time of imaging studies. \citep{prema} restrict their analysis to pregnancies complicated by diabetes mellitus or gestational diabetes mellitus. In contrast, 50-KNR-RF is trained and evaluated on an imbalanced cohort representative of real-world prevalence, without the use of pregnancy outcomes as predictors.

\begin{table*}[!ht]
\centering
\caption{{\bf Classification performance of the meta-model obtained by this work and other models from recent studies.}}
\begin{tabular}{|l||*{3}{c|}}\hline
\backslashbox{Model}{Metric}
&\makebox[4.4em]{\textbf{Accuracy}}&\makebox[4.8em]{\textbf{Sensitivity}}&\makebox[4.4em]{\textbf{Specificity}}\\\hline\hline
\\[-1em]
\cite{esty} & 0.72 & 0.93 & 0.71  \\\hline
\\[-1em]
\cite{esty} & 0.77 & 0.84 & 0.77  \\\hline
\\[-1em]
\cite{wlodarczyk_unet} & 0.78 & 0.74 & 0.85 \\\hline
\\[-1em]
\cite{prema} &  0.76 &  0.84  & 0.73 \\\hline
\\[-1em]
50-KNR-RF & 0.77  &  0.59  & 0.82 \\\hline
\end{tabular}
\label{table:discussion}
\end{table*}

To the best of our knowledge, the only other study on predicting GA at delivery using ML is the one by \cite{heinsalu}, where they investigated models using a simpler version of the pipeline displayed in this work. Their best performing model achieves an $R^2$ score of $0.66$ and a MAE of $1.60$. However, their implementation suffers from data leakage at the imputation stage and was not cross-validated, which makes these results unreliable. Nevertheless, the framework they established is valuable and served as the basis of the present work.

\section{Discussion}

Comprehensive multi-modal fetal data and ML models combine synergistically to predict GA at delivery. The developed pipeline acknowledges and addresses key challenges such as imbalances and missing features in the data set, both of which are common when investigating preterm birth.

Our study provides a proof of concept, but the clinical implementation of a reliable model that could predict the timing of delivery would have important benefits. These include ensuring women are transferred to appropriate neonatal care facilities. A timely transfer helps reduce neonatal mortality and decrease costs \citep{story_antenatal}. Another crucial example is the targeting of therapies to mitigate the effects of prematurity. Specifically, corticosteroids administration can help reduce intra-ventricular haemorrhage and promote lung maturity. The timing of this therapy is highly important, since it works best when administered within a week before delivery, and repeated doses increase the risk of adverse effects such as reduction in birthweight \citep{story_impactuk}. At the current level of performance, the MAE of $2.74$ weeks indicates that predictions are better suited to inform broader antenatal planning than to guide interventions requiring narrow timing windows, such as corticosteroid administration. The $R^{2}$ score of $0.13$ reflects that a substantial proportion of the variance in GA at delivery remains unexplained by the model, and predictions should be used as a complement to current clinical management rather than as a stand-alone tool.

The features selected as the most important are in line with the literature. The importance of cervical length as a predictor in clinical practice is reflected by its consistent use in the models. Placental features obtained from MRI scans were other prominent features, which is in line with the current understanding of the mechanisms leading to iatrogenic preterm birth \citep{hutter, purisch}. The most common clinical indicator, number of previous preterm births, was not a predominant feature.

While our data set could be considered large given the comprehensive data acquisition, including a fetal MR scan in a cohort of women requiring a high level of medical care, its size is an important limitation for ML methods. The few examples of extremely and very preterm subjects available during training help explain the poor performances on these categories. A stratified cross-validation framework was adopted to provide a reliable evaluation of performance and generalisability. Although the MAE remained within the 2-3 weeks range for eight of the ten folds, sensitivity showed substantial variability across folds demonstrating limited ability to detect preterm cases on different data distributions. On the other hand, specificity was higher and more consistent. This is further evidence of the meta-model struggle to overcome strong class imbalance. The variability observed across folds highlights the influence of specific data distribution on performance and underscores the need for further evaluation in larger and more balanced data sets.

The lower performance observed in the subgroup with a previous history of preterm birth (Table \ref{tab:stratified_performance}) further illustrates the challenges imposed by data imbalance. The small number of cases in this group ($n=32$) likely reduced the model’s ability to generalise within this clinically important population. This may reflect both sample size constraints and variability in the underlying mechanisms of recurrent preterm birth.

All MR features were acquired at a single GA for each patient. The biological pathways that lead to preterm birth change progressively across pregnancy \citep{goldenberg}. While deconfounding via internally studentised residuals addresses temporal dependency with respect to GA, the data offer no longitudinal description of the trajectory of the feature values.

Another limitation is the lack of information on the clinical presentation of preterm birth for every patient in the data set. Iatrogenic and spontaneous preterm births have different aetiologies and training separate models for each case could not only yield better predictions, but also help improve the understanding of each clinical presentation by differentiating their most predictive features. Future work will focus on such subgroups and on extracting relevant phenotypes associated with the different types of preterm birth.

Data obtained on a 1.5T and a 0.55T scanner were available for this study. However, these were not included as there is not a straightforward way to extrapolate the signals acquired by scanners with different magnetic field strengths \citep{garcia-eulate}. Future experiments that include these types of data could test the adequacy of the elements of the pipeline, such as the method of internally studentised residuals, to make accurate predictions regardless of field strength. Similarly, as the data used for this paper comprised only singleton pregnancies, the generalisability of the model to multiple gestations remains to be established. Evaluation in twin pregnancies would provide further validation of the robustness of the pipeline across different cohorts.

There are other directions future research can take to expand or improve the methodology presented in this work. The implementation of the models is ready to benefit from larger or more complete data sets. Adding features known for their predictive power, such as quantitative fibronectin measurements, could improve the results. The performance of the meta-model demonstrates that structural and functional information obtained from MRI can be used to predict GA at delivery. An interesting direction is to make predictions directly from the images making use of deep learning techniques, bypassing the problem of missing data and the need of time-consuming measurements made by experienced clinicians. These techniques have been explored to classify preterm and term patients by automatic measurements of cervical length from transvaginal ultrasound \citep{wlodarczyk_unet} and to estimate GA at scan from fetal brain MRI \citep{kojita, shen}.

MRI remains an expensive modality, however, with an increasing use of fetal MRI, the pipeline presented in this study helps to address a question essential for any pregnancy, and can find an application regardless of the indication of the scan. One of the fundamental contributions of this work is that it shows that fetal MR data acquired as part of diagnostic care or research can be used to obtain useful predictions on the GA at delivery, which in turn can inform the care provided to all pregnancies.


\acks{This work was supported by funding from the EPSRC Centre for Doctoral Training in Smart Medical Imaging (EP/S022104/1) to Diego Fajardo-Rojas, from the Wellcome/EPSRC Centre for Medical Engineering (WT203148/Z/16/Z), a UKRI FLF (MR/T018119/1), and DFG Heisenberg funding through the High Tech Agenda Bavaria (502024488) to Jana Hutter, and from the NIHR Advanced Fellowship (NIHR3016640) and the MRC grant (MR/W019469/1) to Lisa Story. 

The authors acknowledge the invaluable help of the radiographers and midwives while acquiring the data presented here.}

%
\ethics{The data used for this work were acquired as part of four ethically approved studies: 14/LO/1169 (Placenta Imaging Project, Fulham Research Ethics Committee, approval received September 23, 2016), 19-SS-0032 (Inflammation study in pregnancy, South East Scotland Ethics Committee, approval  received  March  7,  2019), 21/WA/0075 (Congenital Heart Imaging Programme, Wales Research Ethics Committee, approval received March 8, 2021), and  21/SS/0082 (Individualised Risk prediction of adverse neonatal outcome in pregnancies that deliver preterm using advanced MRI techniques and machine learning, South East Scotland Ethics Committee, approval received March 2022). Informed consent was obtained in all instances.}

\coi{We declare we do not have conflicts of interest.}

\data{All data and code used in this paper are available online at~\url{https://github.com/dfajardorojas/ml-for-preterm-birth-}.}

\bibliography{citation}


\clearpage
\appendix

\onecolumn

\section{Description of the features.}\label{app:features}
Features and outcomes available in the original data set. The first column is the name of the feature, the second column its type (C = continuous, Cat = categorical, D = discrete), the third column provides a short description, and the last column registers how each feature was acquired: clinical background (CB), clinical outcome (CO), structural MRI (sMRI), functional MRI (fMRI), growth ultrasound (GUS), and anomaly ultrasound (AUS).

\tiny
\begin{longtable}[!ht]{|c|c|c|c|} 
    \hline
    \textbf{Feature} & \textbf{Type} & \textbf{Description} & \textbf{Origin}\\ 
    \hline
    \textbf{GA scan ($tag\_ga$)} & C & GA of fetus at time of MR scan & -\\
    \hline
    \textbf{GA ROM ($tag\_garom$)} & C & GA of fetus at time of rupture of the membranes & CO\\
    \hline
    \textbf{Cohort type ($tag\_typ$)} & Cat & Cohort type & CO\\
    \hline
    \textbf{Scanner type ($tag\_scanner$)} & D & 1.5 Tesla, 3 Tesla & -\\
    \hline
    \textbf{Age ($tag\_age$)} & C & Maternal age & CB\\
    \hline
    \textbf{BMI ($tag\_bmi$)} & C & Maternal body mass index & CB\\
    \hline
    \textbf{LOC ($tag\_loc$)} & D & Placental location.  & GU\\
    \hline
    \textbf{GA delivery ($tag\_gadel$)} & C & Gestational age of fetus at delivery & CO\\
    \hline
    \textbf{MOD ($tag\_mod$)} & Cat & Modality of the delivery i.e. C-section. & CO\\
    \hline
    \textbf{Sex ($tag\_sex$)} & Cat & Sex of the fetus (Male, Female) & CO\\
    \hline
    \textbf{BWG ($tag\_bwg$)} & C & Birth weight (grams) & CO\\
    \hline
    \textbf{BWC ($tag\_bwc$)} & C & Birth weight centile & CO\\
    \hline
    \textbf{Parity ($tag\_parity$)} & D & Number of previous pregnancies. & CB\\
    \hline
    \textbf{Parity ($tag\_prev\_ptb$)} & D & Number of previous preterm births. & CB\\
    \hline
    \textbf{BPD ($tag\_bpd$)} & C & Biparietal diameter of the fetal brain & sMRI\\
    \hline
    \textbf{BPD Cent ($tag\_bpd\_cent$)} & C & Biparietal diameter centile of the fetus & sMRI\\
    \hline
    \textbf{TCD ($tag\_ce\_tcd$)} & C & Transcerebral diameter of the fetus & sMRI\\
    \hline
    \textbf{TCD Cent ($tag\_ce\_tcd\_cent$)} & C & Transcerebral diameter centile of the fetus   & sMRI\\
    \hline
    \textbf{Post Hor Diam ($tag\_post\_hor\_diam$)} & C & Diameter of the posterior horn of the fetus  & sMRI\\
    \hline
    \textbf{VOL Body ($tag\_vol\_body$)} & C & Volume of the fetus & sMRI\\
    \hline
    \textbf{Diabetes ($tag\_diabetes$)} & Cat & Diagnosis of diabetes & CB\\
    \hline
    \textbf{HC ($tag\_hc$)} & C & Fetal head circumference at birth & CO\\
    \hline
    \textbf{HC Cent ($tag\_hcc$)} & C & Fetal head circumference centile at birth & CO\\
    \hline
    \textbf{CPTR ($tag\_cptr$)} & C&T2* brain to placenta ratio &  fMRI\\
    \hline
    \textbf{Anom Pi Left ($tag\_anom\_pi\_left$)} & C & Pulsatility index of the left uterine artery & AUS\\
    \hline
    \textbf{Anom Pi Right ($tag\_anom\_pi\_right$)} & C & Pulsatility index of the right uterine artery & AUS\\
    \hline
    \textbf{Anom GA ($tag\_anom\_ga$)} & C & Gestational age at anomaly US & AUS\\
    \hline
    \textbf{Anom LOC ($tag\_anom\_loc$)} & D &  Placental location at anomaly US & AUS\\
    \hline
    \textbf{Anom Cord ($tag\_anom\_cord$)} & D &  Umbilical cord type at anomaly US & AUS\\
    \hline
    \textbf{Anom Cord Ins ($tag\_anom\_cord\_ins$)} & D & Umbilical cord insertion at anomaly US & AUS\\
    \hline
    \textbf{Anom HC ($tag\_anom\_hc$)} & C & Fetal head circumference at anomaly US & AUS\\
    \hline
    \textbf{Anom AC ($tag\_anom\_ac$)} & C & Abdominal circumference at anomaly US & AUS\\
    \hline
    \textbf{Anom BPD ($tag\_anom\_bpd$)} & C & Bi-parietal diameter at anomaly US & AUS\\
     \hline
    \textbf{Anom FL ($tag\_anom\_fl$)} & C & Femur length at anomaly US & GUS\\
    \hline
    \textbf{GU GA ($tag\_gu\_ga$)} & C & Gestational age at growth US & GUS\\
    \hline
    \textbf{GU HC ($tag\_gu\_hc$)} & C & Head circumference at growth US & GUS\\
    \hline
    \textbf{GU AC ($tag\_gu\_ac$)} & C & Abdominal circumference at growth US & GUS\\
    \hline
    \textbf{GU BPD ($tag\_gu\_bpd$)} & C & Bi-parietal diameter at growth US & GUS\\
    \hline
    \textbf{GU FL ($tag\_gu\_fl$)} & C & Femur length at growth US & GUS\\
    \hline
    \textbf{GU Pi ($tag\_gu\_pi$)} & C & Pulsatility index of the umbilical vein & GUS\\
    \hline
    \textbf{GU EFW ($tag\_gu\_efw$)} & C & Estimated foetal weight at growth US & GUS\\
    \hline
    \textbf{GU MCA PI ($tag\_gu\_mca\_pi$)} & C & Pulsatility index of the mid-cerebral artery & GUS\\
    \hline
    \textbf{GU MCA PSV ($tag\_gu\_mca\_psc$)} & C & Peak systolic velocity of the mid-cerebral artery & GUS\\
    \hline
    \textbf{GU MCA CPR ($tag\_gu\_mca\_cpr$)} & C & Cerebroplacental ratio & GUS\\
    \hline
    \textbf{GU Notch ($tag\_gu\_notch$)} & C & Notching present in the uterine artery & GUS\\
    \hline
    \textbf{GU Pi Left ($tag\_gu\_pi\_left$)} & C & Pulsatility index of the left uterine artery & GUS\\
    \hline
    \textbf{GU Pi Right ($tag\_gu\_pi\_right$)} & C & Pulsatility index of the right uterine artery & GUS\\
    \hline
    \textbf{GU EDF ($tag\_gu\_edf$)} & C & End-diastolic flow & GUS\\
    \hline
    \textbf{GU LOC ($tag\_gu\_loc$)} & D & Placental location at growth US & GUS\\
    \hline
    \textbf{GU Cord ($tag\_gu\_cord$)} & D & Umbilical cord type at growth US & GUS\\
    \hline
    \textbf{GU Cord Ins ($tag\_gu\_cord\_ins$)} & C & Umbilical cord insertion at growth US & GUS\\
    \hline
    \textbf{APGAR5 ($tag\_apgar5$)} & C & APGAR score at 5 minutes & CO\\
    \hline
    \textbf{Histo MVM ($tag\_histo\_mvm$)} & C & Histopathology, maternal villi malperfusion & CO\\
    \hline
    \textbf{Histo FVM ($tag\_histo\_fvm$)} & C & Histopathology, foetal villi malperfusion & CO\\
    \hline    
    \textbf{Histo weight ($tag\_histo\_weight$)} & C & Histopathology placental weight & CO\\
    \hline    
    \textbf{Histo chorio ($tag\_histo\_chorio$)} & C & Histopathology chorioamnionitis & CO\\
    \hline    
    \textbf{SMOK ($tag\_smok$)} & D & Maternal smoking status & CB\\
    \hline
    \textbf{IVF ($tag\_ivf$)} & D & In vitro fertilisation (IVF) status & CB\\
    \hline
    \textbf{BWC delivery ($tag\_del\_bwc$)} & C & Birth weight centile (Intergrowth21) at delivery & CO\\
    \hline
    \textbf{Plac T2* mean ($plac\_t2s\_mean$)} & C & Mean whole placental T2* value & fMRI\\
    \hline
    \textbf{Plac T2* vol ($plac\_t2s\_vol$)} & C & Whole placental T2* volume value & fMRI\\
    \hline
    \textbf{Plac T2* lacu ($plac\_t2s\_lacu$)} & C & Placental T2* lacunarity & fMRI\\
    \hline
    \textbf{Plac T2* skew ($plac\_t2s\_skew$)} & C & Placental T2* skewness & fMRI\\
    \hline
    \textbf{Plac T2* kurt ($plac\_t2s\_kurt$)} & C & Placental T2* kurtosis & fMRI\\
    \hline
    \textbf{GU EFW Cent ($tag\_gu\_efw\_cen$)} & C & Expected foetal weigth centile (Intergrowth21) & GUS\\
    \hline
    \textbf{Brain T2* mean ($brain\_t2s\_mean$)} & C & Mean brain T2* value   & fMRI\\
    \hline
    \textbf{Brain T2* vol ($brain\_t2s\_vol$)} & C & Brain T2* volume value   & fMRI\\
    \hline
    \textbf{Brain T2* lacu ($brain\_t2s\_lacu$)} & C & Brain T2* lacunarity & fMRI\\
    \hline
    \textbf{Brain T2* skew ($brain\_t2s\_skew$)} & C & Brain T2* skewness & fMRI\\
    \hline
    \textbf{Brain T2* kurt ($brain\_t2s\_kurt$)} & C & Brain T2* kurtosis & fMRI\\
    \hline
    \textbf{T1 ($t1\_1$)} & C & Mean T1 from MRI & fMRI\\
    \hline
    \textbf{Cervical length ($tag\_cervix\_length$)} & C & Cervical length from sagittal plane  & sMRI\\
    \hline
    \textbf{Cross Study ID ($tag\_complete\_id$)} & C & Anonymous Cross Study Identifier & -\\
    \hline
    \textbf{ Volume eCSF left ($eCSF\_L$)} & C & Volume eCSF left side & sMRI\\
    \hline
    \textbf{ Volume eCSF right ($eCSF\_L$)} & C & Volume eCSF right side & sMRI\\
    \hline
    \textbf{Left cortex ($Cortex\_L$)} & C & Volume cortex left & sMRI\\
    \hline
    \textbf{Right cortex ($Cortex\_L$)} & C & Volume cortex right & sMRI\\
    \hline
    \textbf{Left white matter ($WM\_L$)} & C & White matter volume left & sMRI\\
    \hline
    \textbf{Right white matter ($WM\_R$)} & C & White matter volume right & sMRI\\
    \hline
    \textbf{Left lateral ventricles ($Lat\_ventricle\_L$)} & C & Lateral ventricles volume left & sMRI\\
    \hline
    \textbf{Right lateral ventricles ($Lat\_ventricle\_R$)} & C & Lateral ventricles volume right & sMRI\\
    \hline
    \textbf{Csp volume ($tag\_cervix\_length$)} & C & Cavum septi pellucidi volume & sMRI\\
    \hline
    \textbf{Brainstem volume ($Brainstem$)} & C & Brainstem volume & sMRI\\
    \hline
    \textbf{Left cerebellum volume ($Cerebellum\_L$)} & C & Cerebellum volume left & sMRI\\
    \hline
    \textbf{Right cerebellum volume ($Cerebellum\_R$)} & C & Cerebellum volume right & sMRI\\
    \hline
    \textbf{Vermis volume ($Vermis$)} & C & Vermis volume & sMRI\\
    \hline
    \textbf{Left lentiform volume ($Lentiform\_L$)} & C & Lentiform volume left & sMRI\\
    \hline
    \textbf{Right lentiform volume ($Lentiform\_R$)} & C & Lentiform volume right & sMRI\\
    \hline
    \textbf{Left thalamus volume ($Thalamus\_L$)} & C & Thalamus volume left & sMRI\\
    \hline
    \textbf{Right thalamus volume ($Thalamus\_R$)} & C & Thalamus volume right & sMRI\\
    \hline
    \textbf{Third ventricle volume ($Third\_ventricle$)} & C & Third ventricle volume & sMRI\\
    \hline
    \textbf{Category ($tag\_cat\_norm$)} & D & Assigned category & CO\\
    \hline
    \textbf{Brain volume T2 ($tag\_cervix\_length$)} & C & Complete T2-weighted brain volume & fMRI\\
    \hline\textbf{Blood pressure systole ($tag\_bp\_sys$)} & C & Average systole blood pressure & CB\\
    \hline
    \textbf{Blood pressure diastole ($tag\_bp\_dias$)} & C & Average diastole blood pressure & CB\\
    \hline
    \textbf{Heart rate ($tag\_bp\_hr$)} & C & Average heart rate & CB\\
    \hline
    \textbf{Fetal body volume ($tag\_volume\_wb$)} & C & Whole fetal body volume & sMRI\\
    \hline
    \textbf{Amniotic fluid volume ($tag\_volume\_amniotic$)} & C & Amniotic fluid volume & sMRI\\
    \hline
    \textbf{Cohort at scan ($tag\_control\_at\_scan$)} & C & Cohort assessment at scan & CO\\
    \hline
    \textbf{Mid cerebellar artery ratio ($tag\_gu\_mca\_cpr1$)} & C & Mid cerebellar artery ratio & US\\
    \hline
    \textbf{Adc ($diff\_1$)} & C & Average adc & fMRI\\
    \hline
    \textbf{T2* ($diff\_2$)} & C & Average T2* & fMRI\\
    \hline
    \textbf{T2* perfusion compartment ($diff\_3$)} & C & T2* perfusion compartment & fMRI\\
    \hline
    \textbf{T2* diffusing compartment ($diff\_4$)} & C & T2* diffusing compartment & fMRI\\
    \hline
    \textbf{Adc perfusion compartment ($diff\_5$)} & C & Adc perfusion compartment & fMRI\\
    \hline
    \textbf{Adc diffusing compartment ($diff\_6$)} & C & Adc diffusing compartment & fMRI\\
    \hline
    \textbf{T2* perfusing compartment weighted ($diff\_7$)} & C & T2* perfusing compartment weighted & fMRI\\
    \hline
    \textbf{T2* diffusing compartment weighted ($diff\_8$)} & C & T2* diffusing compartment weighted & fMRI\\
    \hline
    \textbf{Adc perfusion compartment weighted ($diff\_9$)} & C & Adc perfusion compartment weighted & fMRI\\
    \hline
    \textbf{Adc diffusing compartment weighted ($diff\_10$)} & C & Adc diffusing compartment weighted & fMRI\\
    \hline
    \textbf{Perfusion fraction ivim ($diff\_11$)} & C & Perfusion fraction ivim & fMRI\\
    \hline
    \textbf{Left lung T2* mean ($lung\_t2s\_left\_mean$)} & C & Mean left lung T2* value  & fMRI\\
    \hline
    \textbf{Left lung T2* vol ($lung\_t2s\_left\_vol$)} & C & Left lung T2* volume value & fMRI\\
    \hline
    \textbf{Left lung T2* lacu ($lung\_t2s\_left\_lacu$)} & C & Left lung T2* lacunarity & fMRI\\
    \hline
    \textbf{Left lung T2* skew ($lung\_t2s\_left\_skew$)} & C & Left lung T2* skewness & fMRI\\
    \hline
    \textbf{Left lung T2* kurt ($lung\_t2s\_left\_kurt$)} & C & Left lung T2* kurtosis & fMRI\\
    \hline
    \textbf{Right lung T2* mean ($lung\_t2s\_right\_mean$)} & C & Mean right lung T2* value  & fMRI\\
    \hline
    \textbf{Right lung T2* vol ($lung\_t2s\_right\_vol$)} & C & Right lung T2* volume value  & fMRI\\
    \hline
    \textbf{Right lung T2* lacu ($lung\_t2s\_right\_lacu$)} & C & Right lung T2* lacunarity & fMRI\\
    \hline
    \textbf{Right lung T2* skew ($lung\_t2s\_right\_skew$)} & C & Right lung T2* skewness & fMRI\\
    \hline
    \textbf{Right lung T2* kurt ($lung\_t2s\_right\_kurt$)} & C & Right lung T2* kurtosis & fMRI\\
    \hline
    \textbf{Both lungs T2* mean ($lung\_t2s\_both\_mean$)} & C & Mean T2* value of both lungs  & fMRI\\
    \hline
    \textbf{Both lungs T2* vol ($lung\_t2s\_both\_vol$)} & C & Both lungs T2* volume value  & fMRI\\
    \hline    

\end{longtable}

\normalsize

\section{Statistical summary of the features.}\label{app:statistics}
Statistics of the data set after the first preprocessing steps and before imputation.

\footnotesize
\begin{longtable}[!ht] {|l|l|l|l|l|l|}
 \hline
    \textbf{Feature name} & \textbf{Missing} & \textbf{Mean} & \textbf{Median} & \textbf{STD} & \textbf{Skewness} \\ 
    \hline
$ tag\_ga $ & 0 (0.0\%) & 28.18 & 28.14 & 4.56 & -0.11 \\
\hline
$ tag\_typ $ & 23 (9.47\%) & 158.21 & 99.0 & 300.93 & 2.6 \\
\hline
$ tag\_scanner $ & 0 (0.0\%) & 1.0 & 1.0 & 0.0 & 0.0 \\
\hline
$ tag\_age $ & 10 (4.12\%) & 34.18 & 34.3 & 4.7 & -0.27 \\
\hline
$ tag\_bmi $ & 21 (8.64\%) & 23.77 & 23.65 & 2.98 & 0.48 \\
\hline
$ tag\_loc $ & 35 (14.4\%) & 3.53 & 2.0 & 2.09 & 1.76 \\
\hline
$ tag\_gadel $ & 0 (0.0\%) & 37.37 & 38.86 & 4.37 & -1.73 \\
\hline
$ tag\_mod $ & 0 (0.0\%) & 4.34 & 4.0 & 2.48 & 0.14 \\
\hline
$ tag\_sex $ & 8 (3.29\%) & 1.57 & 2.0 & 0.52 & 0.19 \\
\hline
$ tag\_bwg $ & 7 (2.88\%) & 2832.48 & 3062.5 & 917.24 & -0.91 \\
\hline
$ tag\_bwc $ & 107 (44.03\%) & 45.75 & 43.06 & 29.52 & 0.17 \\
\hline
$ tag\_parity $ & 0 (0.0\%) & 0.48 & 0.0 & 0.78 & 2.14 \\
\hline
$ tag\_prev\_ptb $ & 0 (0.0\%) & 0.11 & 0.0 & 0.33 & 2.84 \\
\hline
$ tag\_bpd $ & 122 (50.21\%) & 73.51 & 74.0 & 12.86 & -0.29 \\
\hline
$ tag\_bpd\_cent $ & 124 (51.03\%) & 49.3 & 47.0 & 28.02 & 0.05 \\
\hline
$ tag\_ce\_tcd $ & 123 (50.62\%) & 33.88 & 32.55 & 8.42 & 0.08 \\
\hline
$ tag\_ce\_tcd\_cent $ & 129 (53.09\%) & 51.94 & 52.5 & 23.59 & -0.08 \\
\hline
$ tag\_post\_hor\_diam $ & 125 (51.44\%) & 6.42 & 6.35 & 1.61 & 0.22 \\
\hline
$ eCSF\_L $ & 100 (41.15\%) & 31941.45 & 33049.6 & 11969.29 & 0.17 \\
\hline
$ eCSF\_R $ & 100 (41.15\%) & 30944.74 & 31214.5 & 10941.05 & 0.35 \\
\hline
$ Cortex\_L $ & 100 (41.15\%) & 21351.56 & 16914.0 & 12237.17 & 0.74 \\
\hline
$ Cortex\_R $ & 100 (41.15\%) & 21422.58 & 17092.2 & 11960.04 & 0.73 \\
\hline
$ WM\_L $ & 100 (41.15\%) & 44681.85 & 41335.1 & 20056.3 & 0.35 \\
\hline
$ WM\_R $ & 100 (41.15\%) & 44771.93 & 41127.5 & 20324.06 & 0.37 \\
\hline
$ Lat\_ventricle\_L $ & 100 (41.15\%) & 2154.67 & 2018.12 & 963.81 & 1.01 \\
\hline
$ Lat\_ventricle\_R $ & 100 (41.15\%) & 2008.0 & 1891.96 & 893.35 & 0.89 \\
\hline
$ CSP $ & 100 (41.15\%) & 497.36 & 471.24 & 224.35 & 0.85 \\
\hline
$ Brainstem $ & 100 (41.15\%) & 3514.29 & 3428.88 & 1474.82 & -0.16 \\
\hline
$ Cerebellum\_L $ & 100 (41.15\%) & 2999.35 & 2493.24 & 1914.44 & 0.5 \\
\hline
$ Cerebellum\_R $ & 100 (41.15\%) & 2977.61 & 2348.0 & 1992.55 & 0.47 \\
\hline
$ Vermis $ & 100 (41.15\%) & 936.0 & 831.04 & 535.0 & 0.25 \\
\hline
$ Lentiform\_L $ & 100 (41.15\%) & 2120.58 & 1953.5 & 1059.44 & 0.23 \\
\hline
$ Lentiform\_R $ & 100 (41.15\%) & 2041.67 & 1871.62 & 1033.36 & 0.15 \\
\hline
$ Thalamus\_L $ & 100 (41.15\%) & 1642.67 & 1483.75 & 819.65 & 0.29 \\
\hline
$ Thalamus\_R $ & 100 (41.15\%) & 1590.87 & 1420.96 & 801.94 & 0.26 \\
\hline
$ Third\_ventricle $ & 100 (41.15\%) & 140.56 & 143.68 & 70.51 & -0.08 \\
\hline
$ tag\_diabetes $ & 7 (2.88\%) & 0.5 & 0.0 & 2.13 & 4.3 \\
\hline
$ tag\_cat\_norm $ & 75 (30.86\%) & 1.45 & 1.0 & 0.52 & 0.45 \\
\hline
$ tag\_hc $ & 64 (26.34\%) & 33.3 & 34.0 & 3.03 & -2.23 \\
\hline
$ tag\_hcc $ & 121 (49.79\%) & 53.64 & 56.6 & 33.57 & -0.12 \\
\hline
$ tag\_garom $ & 0 (0.0\%) & 37.0 & 38.86 & 5.07 & -1.78 \\
\hline
$ tag\_gu\_efw\_cen $ & 103 (42.39\%) & 54.11 & 61.19 & 32.35 & -0.41 \\
\hline
$ tag\_vol\_t2w\_complete $ & 100 (41.15\%) & 143758.08 & 134099.98 & 60465.11 & 0.26 \\
\hline
$ tag\_cptr $ & 102 (41.98\%) & 0.38 & 0.36 & 0.12 & 0.93 \\
\hline
$ tag\_histo\_weight $ & 131 (53.91\%) & 417.62 & 453.0 & 130.33 & -0.55 \\
\hline
$ tag\_histo\_mvm $ & 132 (54.32\%) & 0.22 & 0.0 & 0.41 & 1.4 \\
\hline
$ tag\_histo\_fvm $ & 132 (54.32\%) & 0.02 & 0.0 & 0.13 & 7.35 \\
\hline
$ tag\_histo\_chorio $ & 133 (54.73\%) & 0.38 & 0.0 & 0.49 & 0.49 \\
\hline
$ tag\_anom\_pi\_left $ & 191 (78.6\%) & 1.19 & 1.07 & 0.52 & 0.5 \\
\hline
$ tag\_anom\_pi\_right $ & 191 (78.6\%) & 1.22 & 1.12 & 0.56 & 1.19 \\
\hline
$ tag\_anom\_ga $ & 53 (21.81\%) & 20.02 & 20.0 & 0.9 & -3.34 \\
\hline
$ tag\_anom\_loc $ & 35 (14.4\%) & 3.53 & 2.0 & 2.09 & 1.76 \\
\hline
$ tag\_anom\_cord $ & 69 (28.4\%) & 1.11 & 1.0 & 0.56 & 5.32 \\
\hline
$ tag\_anom\_cord\_ins $ & 182 (74.9\%) & 5.7 & 8.0 & 2.86 & -0.66 \\
\hline
$ tag\_anom\_hc $ & 58 (23.87\%) & 172.3 & 171.8 & 9.81 & 1.51 \\
\hline
$ tag\_anom\_ac $ & 58 (23.87\%) & 150.4 & 150.2 & 11.03 & 0.83 \\
\hline
$ tag\_anom\_bpd $ & 71 (29.22\%) & 47.56 & 47.65 & 3.15 & -2.64 \\
\hline
$ tag\_anom\_fl $ & 58 (23.87\%) & 31.72 & 31.7 & 2.53 & 0.23 \\
\hline
$ tag\_gu\_ga $ & 94 (38.68\%) & 28.65 & 28.43 & 4.63 & -0.37 \\
\hline
$ tag\_gu\_hc $ & 98 (40.33\%) & 259.63 & 265.3 & 44.46 & -0.89 \\
\hline
$ tag\_gu\_ac $ & 98 (40.33\%) & 239.14 & 239.3 & 49.19 & -0.41 \\
\hline
$ tag\_gu\_bpd $ & 100 (41.15\%) & 73.0 & 74.8 & 13.2 & -0.72 \\
\hline
$ tag\_gu\_fl $ & 97 (39.92\%) & 52.53 & 53.25 & 10.87 & -0.68 \\
\hline
$ tag\_gu\_pi $ & 112 (46.09\%) & 1.08 & 1.03 & 0.22 & 1.65 \\
\hline
$ tag\_gu\_efw $ & 103 (42.39\%) & 1378.66 & 1229.5 & 727.21 & 0.53 \\
\hline
$ tag\_gu\_mca\_pi $ & 133 (54.73\%) & 1.85 & 1.84 & 0.36 & -0.12 \\
\hline
$ tag\_gu\_mca\_psc $ & 138 (56.79\%) & 43.83 & 42.7 & 13.47 & 0.71 \\
\hline
$ tag\_gu\_mca\_cpr $ & 158 (65.02\%) & 1.82 & 1.83 & 0.45 & 0.0 \\
\hline
$ tag\_gu\_notch $ & 153 (62.96\%) & 1.1 & 1.0 & 0.5 & 5.23 \\
\hline
$ tag\_gu\_pi\_left $ & 153 (62.96\%) & 0.87 & 0.85 & 0.26 & 0.71 \\
\hline
$ tag\_gu\_pi\_right $ & 153 (62.96\%) & 0.87 & 0.86 & 0.26 & 0.91 \\
\hline
$ tag\_gu\_edf $ & 112 (46.09\%) & 1.11 & 1.0 & 0.62 & 6.12 \\
\hline
$ tag\_gu\_loc $ & 110 (45.27\%) & 3.61 & 2.0 & 2.22 & 1.53 \\
\hline
$ tag\_gu\_cord $ & 140 (57.61\%) & 1.06 & 1.0 & 0.44 & 8.1 \\
\hline
$ tag\_gu\_cord\_ins $ & 161 (66.26\%) & 3.28 & 4.0 & 1.28 & -1.24 \\
\hline
$ tag\_apgar5 $ & 26 (10.7\%) & 9.25 & 10.0 & 1.5 & -3.74 \\
\hline
$ tag\_smok $ & 33 (13.58\%) & 1.41 & 1.0 & 1.08 & 2.35 \\
\hline
$ tag\_ivf $ & 25 (10.29\%) & 1.87 & 2.0 & 0.46 & -2.15 \\
\hline
$ tag\_bp\_sys $ & 105 (43.21\%) & 106.22 & 104.22 & 12.78 & 0.54 \\
\hline
$ tag\_bp\_dias $ & 105 (43.21\%) & 65.36 & 64.06 & 10.72 & 0.53 \\
\hline
$ tag\_bp\_hr $ & 118 (48.56\%) & 77.5 & 76.4 & 10.1 & 0.27 \\
\hline
$ tag\_volume\_wb $ & 220 (90.53\%) & 763487.03 & 683058.59 & 313741.75 & 0.59 \\
\hline
$ tag\_volume\_amniotic $ & 220 (90.53\%) & 460971.62 & 526734.02 & 242231.28 & -0.32 \\
\hline
$ tag\_control\_at\_scan $ & 0 (0.0\%) & 0.62 & 1.0 & 0.49 & -0.49 \\
\hline
$ tag\_gu\_mca\_cpr1 $ & 138 (56.79\%) & 1.79 & 1.83 & 0.46 & -0.22 \\
\hline
$ tag\_del\_bwc $ & 7 (2.88\%) & 47.22 & 48.04 & 31.14 & -0.0 \\
\hline
$ plac\_t2s\_mean $ & 0 (0.0\%) & 57.92 & 59.69 & 19.11 & -0.01 \\
\hline
$ plac\_t2s\_vol $ & 0 (0.0\%) & 366.89 & 340.31 & 173.32 & 0.77 \\
\hline
$ plac\_t2s\_lacu $ & 0 (0.0\%) & 19.47 & 19.08 & 6.16 & 2.53 \\
\hline
$ plac\_t2s\_skew $ & 0 (0.0\%) & 13.33 & 2.36 & 30.53 & 3.49 \\
\hline
$ plac\_t2s\_kurt $ & 0 (0.0\%) & 1.34 & 0.68 & 1.89 & 2.37 \\
\hline
$ lung\_t2s\_left\_mean $ & 159 (65.43\%) & 65.26 & 62.16 & 18.45 & 1.21 \\
\hline
$ lung\_t2s\_left\_vol $ & 159 (65.43\%) & 13.88 & 11.08 & 8.71 & 1.4 \\
\hline
$ lung\_t2s\_left\_lacu $ & 159 (65.43\%) & 20.06 & 18.43 & 8.72 & 1.58 \\
\hline
$ lung\_t2s\_left\_skew $ & 159 (65.43\%) & 7.27 & 1.78 & 13.54 & 2.99 \\
\hline
$ lung\_t2s\_left\_kurt $ & 159 (65.43\%) & 0.99 & 0.58 & 1.16 & 1.81 \\
\hline
$ lung\_t2s\_both\_mean $ & 155 (63.79\%) & 67.06 & 62.62 & 19.26 & 1.17 \\
\hline
$ lung\_t2s\_both\_vol $ & 155 (63.79\%) & 31.95 & 26.15 & 20.13 & 1.2 \\
\hline
$ tag\_vol\_body $ & 100 (41.15\%) & 1009.95 & 971.36 & 456.13 & 8.94 \\
\hline
$ tag\_cervix\_length $ & 26 (10.7\%) & 30.7 & 31.48 & 10.39 & -0.71 \\
\hline
$ brain\_t2s\_mean $ & 101 (41.56\%) & 165.0 & 174.29 & 42.69 & -0.56 \\
\hline
$ brain\_t2s\_vol $ & 101 (41.56\%) & 128.08 & 109.09 & 72.89 & 1.32 \\
\hline
$ brain\_t2s\_lacu $ & 101 (41.56\%) & 68.79 & 73.92 & 19.64 & -0.77 \\
\hline
$ brain\_t2s\_skew $ & 101 (41.56\%) & 2.27 & 0.17 & 6.07 & 4.43 \\
\hline
$ brain\_t2s\_kurt $ & 101 (41.56\%) & 0.91 & 0.73 & 0.67 & 1.56 \\
\hline
$ lung\_t2s\_right\_mean $ & 158 (65.02\%) & 67.97 & 63.24 & 19.56 & 1.07 \\
\hline
$ lung\_t2s\_right\_vol $ & 158 (65.02\%) & 19.64 & 15.85 & 13.1 & 1.15 \\
\hline
$ lung\_t2s\_right\_lacu $ & 158 (65.02\%) & 22.13 & 19.95 & 9.71 & 1.22 \\
\hline
$ lung\_t2s\_right\_skew $ & 158 (65.02\%) & 7.66 & 2.42 & 14.4 & 4.1 \\
\hline
$ lung\_t2s\_right\_kurt $ & 158 (65.02\%) & 1.14 & 0.82 & 1.19 & 2.08 \\
\hline
$ t1\_1 $ & 199 (81.89\%) & 1003.34 & 1016.46 & 243.97 & -0.8 \\
\hline
$ diff\_1 $ & 140 (57.61\%) & 55.75 & 57.47 & 19.34 & -0.22 \\
\hline
$ diff\_2 $ & 140 (57.61\%) & 0.0 & 0.0 & 0.0 & 1.14 \\
\hline
$ diff\_3 $ & 140 (57.61\%) & 0.0 & 0.0 & 0.0 & 2.41 \\
\hline
$ diff\_4 $ & 140 (57.61\%) & 0.0 & 0.0 & 0.0 & 0.0 \\
\hline
$ diff\_5 $ & 140 (57.61\%) & 0.01 & 0.0 & 0.01 & 3.3 \\
\hline
$ diff\_6 $ & 140 (57.61\%) & 0.0 & 0.0 & 0.0 & 0.0 \\
\hline
$ diff\_7 $ & 140 (57.61\%) & 52.51 & 52.21 & 32.07 & 0.11 \\
\hline
$ diff\_8 $ & 140 (57.61\%) & 0.03 & 0.03 & 0.01 & 1.02 \\
\hline
$ diff\_9 $ & 140 (57.61\%) & 76.01 & 75.25 & 22.84 & 0.99 \\
\hline
$ diff\_10 $ & 140 (57.61\%) & 0.0 & 0.0 & 0.0 & 0.62 \\
\hline
$ diff\_11 $ & 140 (57.61\%) & 0.33 & 0.3 & 0.17 & 0.68 \\
\hline
$ tag\_complete\_id $ & 0 (0.0\%) & 2400682.76 & 1000187.0 & 2864221.51 & 1.98 \\
\hline
    
\end{longtable}

\normalsize

\section{Hyperparameters of the base models.}\label{app:hyperparameters}
Hyperparameters investigated by the grid search for each of the base models

\begin{table}[H]
\tiny
    \centering
    \begin{tabular}{|c|c|c|c|c|c|}
    \hline
    \multicolumn{2}{|c|}{\textbf{Random Forests}} & \multicolumn{2}{|c|}{\textbf{Support Vector Regression}} & \multicolumn{2}{|c|}{\textbf{XGBoost}} \\
    \hline
    \textbf{Hyperparameter} & \textbf{Values} & \textbf{Hyperparameter} & \textbf{Values}  & \textbf{Hyperparameter} & \textbf{Values} \\
    \hline
     $\boldsymbol {max\_depth}$  & [3, 5, 10, 20, 50, 100] & $\boldsymbol {C}$ & [0.001, 0.01, 0.1, 1, 10, 100, 1000] & $\boldsymbol {max\_depth}$  & [3, 5, 7, 10, 20, 50] \\
    \hline
    $\boldsymbol {max\_features}$ & ['auto', 'sqrt', 'log2'] & $\boldsymbol{gamma}$ & ['scale', 'auto'] & $\boldsymbol {learning\_rate}$  & [0.01, 0.05, 0.1, 0.3, 0.5]  \\
    \hline
    $\boldsymbol {n\_estimators} $ & [5, 10, 20, 50, 100, 250] & $\boldsymbol {kernel}$ & ['rbf', 'poly', 'sigmoid', 'linear'] & $\boldsymbol {min\_child\_weight}$  & [1,3,5,7]  \\
    \hline
    &  &  $\boldsymbol{epsilon}$ & [0.001 ,0.01 ,0.1, 0.5, 1] & $\boldsymbol{gamma}$ & [0.1, 0.5, 0.8, 2, 5, 10]    \\
    \hline
    &  &  $\boldsymbol {degree}$ & [2, 3] & $\boldsymbol {colsample\_bytree}$  & [0.3, 0.5, 0.7]  \\
    \hline
    \end{tabular}
\end{table}

\section{Details of First Preprocessing Steps.}\label{app:figure1}
Number of subjects kept after each initial preprocessing step.

\begin{figure}[H]
    \centering
    \includegraphics[width=5cm]{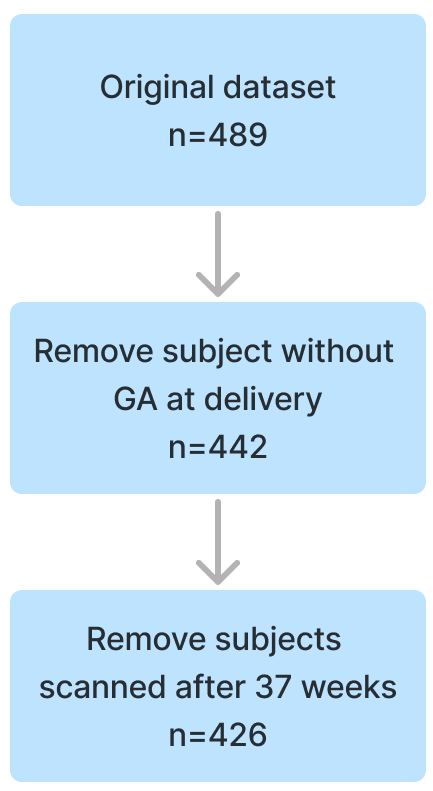}
    \label{}
\end{figure}

\section{Pseudocode of the MICE algorithm.}\label{app:mice}

\begin{algorithm}[H]
\caption{Pseudocode of the MICE algorithm}

\hspace*{\algorithmicindent} \textbf{Input:} Data matrix $D_{ij}=(d_{ij})$, Number of iterations MaxIter.

\hspace*{\algorithmicindent} \textbf{Output:} Imputed matrix $\hat{D}_{ij}$.

\begin{algorithmic}[1]

    \State Impute \(D_{ij}\) with column means: \(\hat{D}_{ij} = (\hat{d_{ij}}) \gets \text{FillWithMeans}(D_{ij})\)
    \State \(k \gets 1\)
    \While{\(k \leq \text{MaxIter}\)}
        \For{each column \(D_{j}\) in \(D_{ij}\)}
                \State Set imputations for \(D_{j}\) to missing in \(\hat{D}_{ij}\)
                \State \(f_{j} \gets \text{FitRegressionModel} (\hat{D}_{ij}, \hat{D}_{j})\)
                \For{each row \(D_{i}\) in \(D_{ij}\)}
                \If{\(\hat{d}_{ij}\) is missing in \(\hat{D}_{ij}\)}
                \State Impute missing value: \(\hat{d_{ij}} \gets\) \(f_{j} (\hat{D}_{i}\setminus\{\hat{d_{ij}}\})\)
                \EndIf
                \EndFor
        \EndFor
        \State \(k \gets k + 1\)
    \EndWhile
    \State \textbf{return} \(\hat{D}_{ij}\) 

\end{algorithmic}
\end{algorithm}

\end{document}